\documentclass[twocolumn,preprintnumbers,amsmath,amssymb]{revtex4}
\usepackage{graphicx}
\usepackage{amssymb}

\usepackage{multirow}

\begin{document}

\title{Collaboration between parallel connected neural networks - A possible
criterion for distinguishing artificial neural networks from natural organs}
\author{Guang Ping He}
\email{hegp@mail.sysu.edu.cn}
\affiliation{School of Physics, Sun Yat-sen University, Guangzhou 510275, China}

\begin{abstract}
We find experimentally that when artificial neural networks are connected in parallel and trained together, they display the following properties. (i)
When the parallel-connected neural network (PNN) is optimized, each
sub-network in the connection is not optimized. (ii) The contribution of an
inferior sub-network to the whole PNN can be on par with that of the superior
sub-network. (iii) The PNN can output the correct result even when all
sub-networks give incorrect results. These properties are unlikely for
natural biological sense organs. Therefore, they could serve as a simple yet effective
criterion for measuring the bionic level of neural networks. With this
criterion, we further show that when serving as the activation function, the
ReLU function can make an artificial neural network more bionic than the
sigmoid and Tanh functions do.
\end{abstract}

\maketitle



When we see Michael Jackson on TV, it seems impossible that our left eye
takes him as John, our right eye treats him as Paul, while our brain
recognizes him as Michael correctly. Similarly, we know that there is smoke
when we see the smoke and smell the smoke. It is less likely that our eyes
see cola while our nose smells vinegar and then our brain concludes that it is
smoke. From our own life experience, when two sense organs (e.g., eyes)
share the same input (e.g., the image on the screen) and output (e.g., the
brain) without direct connection in the middle, each of them
would give the same judgement when the final output is correct.

But what if two artificial neural networks are connected in parallel and
trained together? That is, they share the same input and output layers,
while their hidden layers are separated from each other. Surely, we all know
that even the state-of-the-art artificial neural networks do not replicate
biological neural networks faithfully, but little is known about what
exactly is the difference, especially in a quantitative way \cite{ml106,ml105}. In this work, by using the classification of
Modified National Institute of Standards and Technology (MNIST) data set as
an example, we find some properties that clearly distinguish the
parallel-connected artificial neural networks (PNNs) apart from natural sense
organs. Moreover, different types of PNNs will show these divergence to a
different level. This finding could serve as a criterion for measuring the
bionic level of an artificial neural network.

\section*{Results}

\textbf{Parallel connections of neural networks}

Fig.1 shows the most basic structure of a conventional neural network, where
each neuron connects with every neuron in the neighboring layers. In the
following we will call it a fully-connected neural network (FNN), and use
the notation $[n_{0},n_{1},...,n_{m}]$ to denote an $m$-layer (without
counting the input layer) FNN, where $n_{0},$ $n_{i}$ ($i=1,...,m-1$) and $%
n_{m}$ are the numbers of neurons in the input layer, the $i$th hidden layer
and the output layer, respectively.


\begin{figure}[htbp]
\includegraphics[scale=0.85]{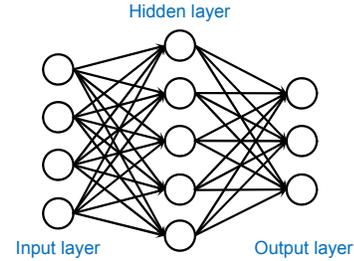}
\caption{A $[4,5,3]$ fully-connected neural network (FNN), where $4$, $5$ and $%
3 $ denote the numbers of neurons in the input, hidden and
output layers, respectively.}
\label{fig:epsart}
\end{figure}


In this work, however, we are interested in PNNs, which contain two or more
FNNs (called sub-networks) being connected in parallel by sharing the
same input and output layers, while there is no connection between the
neurons in the hidden layers of different sub-networks, as illustrated in
Fig.2. Let $[n_{0},n_{1},...,n_{m_{1}}]+[n_{0}^{\prime
},n_{1}^{\prime },...,n_{m_{2}}^{\prime }]$\ denote a PNN consisting of two sub-networks $%
[n_{0},n_{1},...,n_{m_{1}}]$\ and $[n_{0}^{\prime },n_{1}^{\prime
},...,n_{m_{2}}^{\prime }]$\ (surely there should be $%
n_{0}=n_{0}^{\prime }$\ and $n_{m_{1}}=n_{m_{2}}^{\prime }$).


\begin{figure}[htbp]
\includegraphics[scale=0.85]{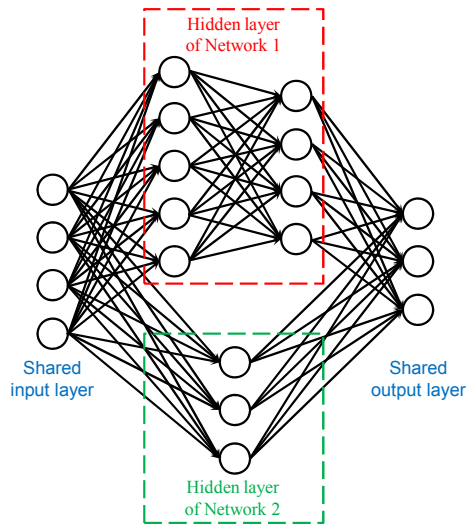}
\caption{A $[4,5,4,3]+[4,3,3]$ parallel-connected neural network (PNN), where
Network 1 is a $[4,5,4,3]$\ FNN and Network 2 is a $[4,3,3]$ FNN.}
\label{fig:epsart}
\end{figure}


PNN should not be confused with the technique where two or more sub-networks
run in parallel and then the final result is obtained via majority voting.
In this technique, the sub-networks are trained and optimized separately.
But in a PNN, they are connected via the output layer and trained together,
so that they will interact with each other and the optimization result will
be different (as we shall show below). PNN also differs from the pooling
technique used in, e.g., AlexNet \cite%
{ml100} because: (i) A network with pooling can be viewed as a special case
of an FNN where some of the weight parameters are set to $0$, while this is
not the case of a PNN consisting of sub-networks with different numbers of
layers. (ii) When the pooling technique is applied to the first hidden
layer, some neurons in this layer connect with only a portion of the neurons
in the input layer. But for every sub-network in a PNN, each neuron in the
first hidden layer connects with all neurons in the input layer.

Note that the parallel connection of several 2-layer FNNs is actually still
a single (but larger) FNN. For example, the $[4,5,3]$ FNN showed in Fig.1
can also be viewed as a $[4,3,3]+[4,2,3]$\ PNN. Therefore, here we study only the PNNs which contain at least one $m$-layer FNN with $m\geq 3$.

\bigskip

\textbf{Experiments}

The MNIST data set \cite{MNIST} is a widely-used resource for machine
learning research \cite{MNIST2}. It contains $70000$ greyscale $28\times 28$
pixel images of handwritten digits $0\symbol{126}9$. Here we also use neural
networks as classifiers to recognize the digits in these images, and study
the classification accuracies of the FNNs within the PNNs. For this purpose,
the numbers of neurons in the input and output layers of each of our PNNs and
FNNs are taken to be $28^{2}=784$\ and $10$, respectively, corresponding to
the $784$ input pixels and the $10$ possible output digits $0\symbol{126}9$.

We conducted the following experiments (see \textit{Methods} section for the
details of the computer programs), each of which displays a type of divergence from the properties of natural sense organs.

\bigskip

\textit{(1) Evolution of the classification accuracies.}

(1.1) The $[784,48,35,10]+[784,50,10]$ PNN. We choose these values because
the total number of learnable parameters (biases and weights) of the $%
[784,48,35,10]$ FNN is $(48+35+10)+(784\times 48+48\times 35+35\times
10)=39755$, while that of the $[784,50,10]$ FNN is $(50+10)+(784\times
50+50\times 10)=39760$.\ From past experience, two FNNs with similar numbers
of parameters will generally have similar performance when working alone.
Thus, it will be interesting to see how they perform when being
parallel-connected.

The activation function of every neuron is chosen as the sigmoid function
(except in experiment (4) below)%
\begin{equation}
f(z)=\frac{1}{1+e^{-z}},
\end{equation}%
with%
\begin{equation}
z=w\cdot x+b
\end{equation}%
denoting the weighted input to the neuron, where $w$ is the weight matrix, $%
b $ is the bias vector, and $x$ is the output from the previous layer.

Let us refer the $[784,48,35,10]$ FNN and the $[784,50,10]$ FNN as Network 1
and Network 2, respectively. We first trained them separately for $60$
epochs (i.e., epochs $0\symbol{126}59$), then connected them in parallel and
trained together for another $40$ epochs (i.e., epochs $60\symbol{126}99$).
In the following, we call this procedure \textit{training method A}. The
classification accuracies of each sub-network and the whole PNN in each
epoch is shown in Fig.3.

Note that there are two ways to calculate the classification accuracy of a
single sub-network within a PNN. Before being connected with the other
sub-network, each sub-network has its own output layer. For Network $i$ ($%
i=1,2$), the weighted input to an output neuron (i.e., the neuron in the
output layer) has the form%
\begin{equation}
z_{i}=w_{i}\cdot x_{i}+b_{i}.  \label{NN}
\end{equation}%
At this time, there is no doubt that the classification accuracy of Network $%
i$ is calculated by taking $z_{i}$\ into the sigmoid function to obtain the
activation value of the neuron, then picking out the output
neuron having the maximal activation value, and comparing its label with
the actual handwritten digit corresponding to this input.

But when the two sub-networks are connected, they share the same output
layer. The total weighted input to an output neuron becomes%
\begin{equation}
z=(w_{1}\cdot x_{1}+w_{2}\cdot x_{2})+b
\end{equation}%
where $b=b_{1}+b_{2}$. If we want to calculate the current classification
accuracy of Network $i$ in the PNN, it is natural to set $w_{i^{\prime }}=0$
to bypass the contribution of Network $i^{\prime }$\ ($i^{\prime }\neq i$%
). But should we also set $b_{i^{\prime }}=0$ and use $b_{i}$ alone to
calculate $z$? That is, should we still use Eq. (\ref{NN}), or use%
\begin{equation}
z=w_{i}\cdot x_{i}+b  \label{NNp}
\end{equation}%
instead?

On one hand, for a fair comparison with the case when Network $i$ was
trained separately, it seems that we should still use $z=w_{i}\cdot
x_{i}+b_{i}$ to calculate the activation value and get the classification
accuracy. Let $\alpha _{i}$ denote the accuracy achieved this way.

On the other hand, however, after the connection is made, the PNN is trained
and optimized with $b=b_{1}+b_{2}$\ being treated as an entirety that serves
as the bias value inherent in the output neuron. Therefore, using $z=w_{i}\cdot x_{i}+b$ to calculate the classification
accuracy of Network $i$ appears to be a better description of its contribution
within the PNN when Network $i^{\prime }$\ ($i^{\prime }\neq i$) is
bypassed. Let $\alpha _{i}^{\prime }$ denote the accuracy thus achieved.

To avoid the dilemma, in Fig.3 we show both $\alpha _{1}$, $\alpha _{2}$\
and $\alpha _{1}^{\prime }$, $\alpha _{2}^{\prime }$. Fortunately, though
their exact values are different, we shall see that they display the same
qualitative results. In epochs $0\symbol{126}59$ where both Networks 1 and 2
are trained separately, Fig.3 shows that $\alpha _{1}$ and $\alpha _{2}$\
grow rapidly in the first few epochs, then tend to saturate. There is always
$\alpha _{1}\simeq \alpha _{2}$. More rigorously, we find that their
optimized values are%
\begin{equation}
\max (\alpha _{1})=97.63\%
\end{equation}%
at epoch $50$\ and%
\begin{equation}
\max (\alpha _{2})=97.49\%
\end{equation}%
at epoch $56$, respectively. This agrees with our guess that two FNNs with
similar numbers of parameters generally have similar performance. (Note that
$\alpha _{1}^{\prime }$, $\alpha _{2}^{\prime }$\ and the classification
accuracy $\alpha _{para}$ of the whole PNN\ have no actual meanings at this
stage since the two sub-networks have not been connected yet. Still we plot
them in all figures for reference.)


\begin{figure}[tbp]
\includegraphics[scale=0.7]{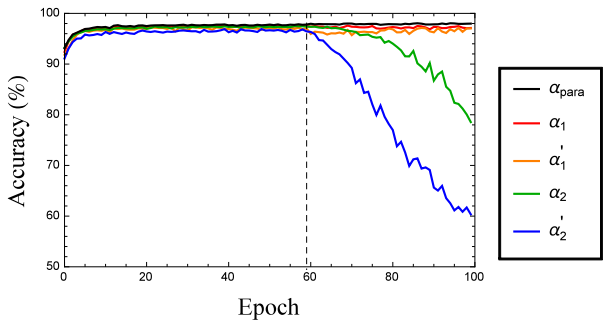}
\caption{The classification accuracy of the $[784,48,35,10]+[784,50,10]$ PNN
during $100$ epochs of training. The two sub-networks are trained separately
in epochs $0\symbol{126}59$, then connected in parallel and trained together
in epochs $60\symbol{126}99$. The activation function of each neuron is
chosen as the sigmoid function. $\alpha _{para}$ denotes the classification
accuracy of the whole PNN. $\alpha _{1}$ ($\alpha _{2}$) is the accuracy of
Network 1 (Network 2) calculated from Eq. (\ref{NN}), and $\alpha
_{1}^{\prime }$ ($\alpha _{2}^{\prime }$) is the accuracy of Network 1
(Network 2) calculated from Eq. (\ref{NNp}).}
\label{fig:epsart}
\end{figure}


But an interesting change emerges when Networks 1 and 2 are connected at
epoch $60$. From Fig.3 we can see that the performance of Network 1 (either
measured by $\alpha _{1}$ or $\alpha _{1}^{\prime }$) remains much the same,
but $\alpha _{2}$\ and $\alpha _{2}^{\prime }$ drop dramatically when the
connected PNN is trained as a whole, showing that Network 2 alone gives less
and less correct classification results in the PNN. At epoch $99$ we have $%
\alpha _{2}=78.55\%$\ and $\alpha _{2}^{\prime }=60.35\%$, both are
significantly lower than the maximal accuracy $\max (\alpha _{2})$ that Network 2
could reach alone. In\ fact, detailed data (see \textit{Data availability} section)
shows that $\alpha _{1}$ and $\alpha _{1}^{\prime }$ drop too, though not by
much. Even so, the whole PNN outperforms each single FNN with its
classification accuracy $\alpha _{para}$ surpassing both $\max (\alpha _{1})$
and $\max (\alpha _{2})$. At epoch $86$ it is optimized with%
\begin{equation}
\max (\alpha _{para})=98.05\%,
\end{equation}%
where we have%
\begin{eqnarray}
\alpha _{1} &=&97.51\%,  \nonumber \\
\alpha _{1}^{\prime } &=&96.98\%,  \nonumber \\
\alpha _{2} &=&90.14\%,  \nonumber \\
\alpha _{2}^{\prime } &=&71.39\%.
\end{eqnarray}

These results indicate that in a PNN, the sub-networks work in a different
way than they do alone. It is distinct from the majority voting scheme where
each sub-network does its best to find the correct results to make the
combination of sub-networks do even better. On the contrary, it seems that
once connected, the PNN can automatically divide the labor of the
sub-networks, so that at least some (if not all) of the sub-networks (e.g.,
Network 2 in the current case) turn to capture other features of the input,
instead of letting every sub-network to reach directly for the solutions of
the classification task alone (we will elaborate this point further in (3)
below). Consequently, when the whole PNN is optimized, the sub-networks tend
to work on a status that appears to be unoptimized on their own.

(1.2) Now let us train the $[784,48,35,10]+[784,50,10]$ PNN for $100$ epochs
with the two sub-networks connected in parallel from the very start. We call
this procedure \textit{training method B} thereafter. The result is shown
in Fig.4.


\begin{figure}[b]
\includegraphics[scale=0.7]{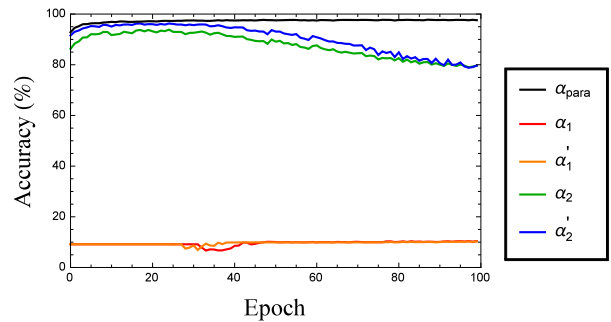}
\caption{The classification accuracy of the $[784,48,35,10]+[784,50,10]$ PNN
during $100$ epochs of training. The two sub-networks are connected and
trained together in all the $100$ epochs. The notations are the same as
those of Fig.3.}
\label{fig:epsart}
\end{figure}


We can see that the property found in experiment (1.1) becomes more
significant. While the classification accuracy $\alpha _{para}$ of the whole
PNN remains high, both $\alpha _{1}^{\prime }$ ($\simeq \alpha _{1}$) and $%
\alpha _{2}^{\prime }$ ($\simeq \alpha _{2}$) are much lower than either $%
\alpha _{para}$ or the maximums that they could reach alone (i.e., $\max
(\alpha _{1})=97.63\%$\ and $\max (\alpha _{2})=97.49\%$ as found in (1.1)).
When $\alpha _{para}$ peaks at epoch $87$ with%
\begin{equation}
\max (\alpha _{para})=97.77\%,
\end{equation}%
there are%
\begin{eqnarray}
\alpha _{1} &=&10.18\%,  \nonumber \\
\alpha _{1}^{\prime } &=&10.06\%,  \nonumber \\
\alpha _{2} &=&81.08\%,  \nonumber \\
\alpha _{2}^{\prime } &=&82.27\%.
\end{eqnarray}%
Thus, it shows that the trend of dividing the labor within the PNN turns out
to be more obvious when the sub-networks are connected earlier and longer.

Another significant observation is that we have $\alpha _{1}^{\prime
}<<\alpha _{2}^{\prime }$\ now, in contrast to $\alpha _{1}^{\prime
}>>\alpha _{2}^{\prime }$\ found in (1.1). It means that the $[784,48,35,10]$
sub-network is not necessarily superior to the $[784,50,10]$ one, nor vice
versa. The division of labor is determined more by the way how the biases
and weights of the sub-networks are initialized, rather than the structure
(e.g., the number of layers) of the sub-networks.

(1.3) The $[784,48,35,10]+[784,48,35,10]$ PNN. The above $%
[784,48,35,10]+[784,50,10]$ PNN is the combination of two different types of
FNNs. We may take them loosely as an eye and a nose. Now let us study the
combination of two eyes.

Again, we first trained them using training method A. The classification
accuracies are shown in Fig.5. The trend in (1.1) also appears here, but
less significant (note that the $y$-coordinate in Fig.5 starts from $90\% $, while it is $50\% $ in Fig.3). More precisely, when being trained separately, the two
sub-networks are optimized at epoch $56$ simultaneously, where they reach%
\begin{equation}
\max (\alpha _{1})=97.81\%
\end{equation}%
and%
\begin{equation}
\max (\alpha _{2})=97.63\%,
\end{equation}%
respectively. But when they are connected at epoch $60$, $%
\alpha _{1}$, $\alpha _{1}^{\prime }$, $\alpha _{2}$, and $\alpha
_{2}^{\prime }$ all start to drop. When the whole PNN is found optimized at epoch $%
90$, there are%
\begin{eqnarray}
\max (\alpha _{para}) &=&98.17\%,  \nonumber \\
\alpha _{1} &=&96.22\%,  \nonumber \\
\alpha _{1}^{\prime } &=&94.95\%,  \nonumber \\
\alpha _{2} &=&96.53\%,  \nonumber \\
\alpha _{2}^{\prime } &=&96.17\%.
\end{eqnarray}%
At epoch $99$ when the training ends, we have%
\begin{eqnarray}
\alpha _{para} &=&98.13\%,  \nonumber \\
\alpha _{1} &=&94.35\%,  \nonumber \\
\alpha _{1}^{\prime } &=&93.70\%,  \nonumber \\
\alpha _{2} &=&96.31\%,  \nonumber \\
\alpha _{2}^{\prime } &=&95.40\%.
\end{eqnarray}


\begin{figure}[htbp]
\includegraphics[scale=0.7]{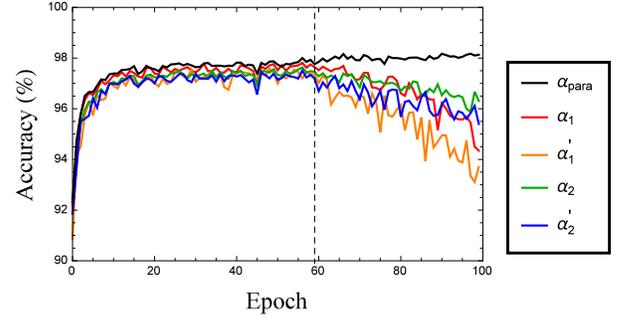}
\caption{The classification accuracy of the $[784,48,35,10]+[784,48,35,10]$ PNN
during $100$ epochs of training. The two sub-networks are trained separately
in epochs $0\symbol{126}59$, then connected in parallel and trained together
in epochs $60\symbol{126}99$. The notations are the same as those of Fig.3.}
\label{fig:epsart}
\end{figure}


At first glance this result looks \textquotedblleft
normal\textquotedblright\ and more like the real combination of two eyes
than the previous two experiments, since $\alpha _{1}$
and $\alpha _{2}$ do not stray away from their optimized values too much after being connected.
But wait! In the current case the two sub-networks are trained separately
for $60$ epochs first. For natural sense organs this will be very unusual\
instead. (When was the last time that you saw a human infant growing with two
eyes being trained separately?) Therefore, to compare with the behavior of
natural sense organs, the following experiment should be more suitable.

(1.4) Like (1.2), we also trained the $[784,48,35,10]+[784,48,35,10]$ PNN
using training method B. The result is shown in Fig.6.


\begin{figure}[b]
\includegraphics[scale=0.7]{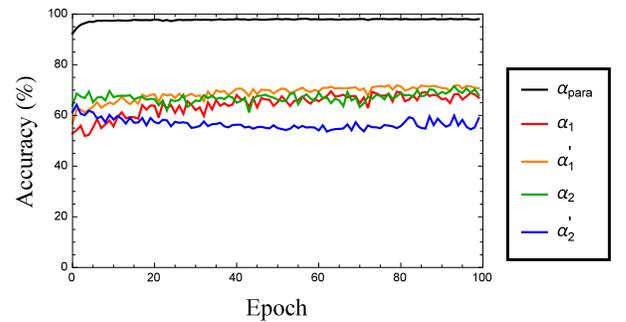}
\caption{The classification accuracy of the $[784,48,35,10]+[784,48,35,10]$ PNN
during $100$ epochs of training. The two sub-networks are connected and
trained together in all the $100$ epochs. The notations are the same as
those of Fig.3.}
\label{fig:epsart}
\end{figure}


We can see that the dropping of $\alpha _{1}$, $\alpha _{1}^{\prime }$, $%
\alpha _{2}$, and $\alpha _{2}^{\prime }$ becomes obvious again (comparing
with the maximal values that they could reach alone in the first $60$\
epochs of (1.1) and (1.3)). When the PNN is optimized at epoch $57$ with%
\begin{equation}
\max (\alpha _{para})=98.16\%,
\end{equation}%
there are%
\begin{eqnarray}
\alpha _{1} &=&65.82\%,  \nonumber \\
\alpha _{1}^{\prime } &=&69.77\%,  \nonumber \\
\alpha _{2} &=&68.32\%,  \nonumber \\
\alpha _{2}^{\prime } &=&55.22\%.
\end{eqnarray}

In brief, all the above experiments display a property different from natural
sense organs. That is, as mentioned at the end of (1.1), when two artificial
neural networks are connected in parallel and trained together as a PNN,
each of them tends to work on an unoptimized status when the whole PNN is
optimized. On the contrary, despite that there are still many unknown
mysteries about life, at least we can be sure that barely anyone has experienced instant
deterioration of the eye sight and the sense of smell though our eyes and
nose are connected to the same brain and fed with the same input.

\bigskip

\textit{(2) Weight distribution.}

Here we study a $[784,48,35,10]+[784,24,18,10]$ PNN using training method B.
In this PNN, both sub-networks have two hidden layers. But the numbers of
neurons and parameters of Network 2 are approximately 50\% of those of
Network 1, so it is expected to be less powerful if working alone. Now let
us study whether it also plays less contribution in the PNN.

The classification accuracies are shown in Fig.7. Like Fig.4 and Fig.6, $%
\alpha _{para}$ remains high while $\alpha _{1}$, $\alpha _{1}^{\prime }$, $%
\alpha _{2}$, and $\alpha _{2}^{\prime }$ are much lower. At epoch $89$ the
PNN is optimized with%
\begin{equation}
\max (\alpha _{para})=98.02\%,
\end{equation}%
where%
\begin{eqnarray}
\alpha _{1} &=&76.08\%,  \nonumber \\
\alpha _{1}^{\prime } &=&75.17\%,  \nonumber \\
\alpha _{2} &=&38.33\%,  \nonumber \\
\alpha _{2}^{\prime } &=&43.98\%.
\end{eqnarray}%
It is not surprising to find $\alpha _{1}>>\alpha _{2}$\ and $\alpha
_{1}^{\prime }>>\alpha _{2}^{\prime }$.


\begin{figure}[htbp]
\includegraphics[scale=0.7]{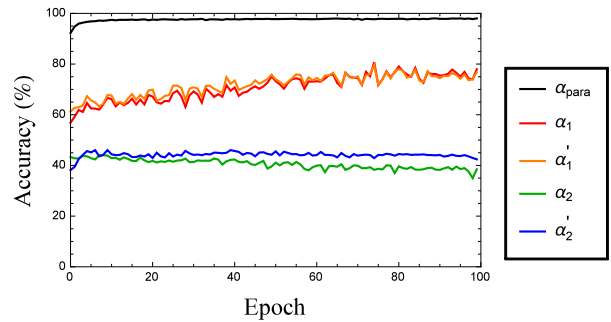}
\caption{The classification accuracy of the $[784,48,35,10]+[784,24,18,10]$ PNN
during $100$ epochs of training. The two sub-networks are connected and
trained together in all the $100$ epochs. The notations are the same as
those of Fig.3.}
\label{fig:epsart}
\end{figure}


But a smaller $\alpha _{i}$\ ($\alpha _{i}^{\prime }$) does not \textit{%
always} mean that Network $i$ contributes less in the PNN (although
sometimes it does). A better measure is the weights at the output layer. If
the weights corresponding to one of the sub-network is much smaller than
these of the other, then it means that the output neurons actually ignore
the inputs from the first sub-network, so that the PNN leans mainly on the
latter to produce the classification output.

When the current PNN is optimized at epoch $89$, we extract the weights at
the output layer (more rigorously, the weights of the outputs of the neurons
in the last hidden layer which are used for calculating the weighted inputs
to the output neurons). The data is shown in Fig.8. Intriguingly, we can see
that the weights of Networks 1 and 2 are on the same level, even though $%
\alpha _{1}$ ($\alpha _{1}^{\prime }$) is almost twice as much as $\alpha
_{2}$ ($\alpha _{2}^{\prime }$).


\begin{figure}[htbp]
\includegraphics[scale=0.7]{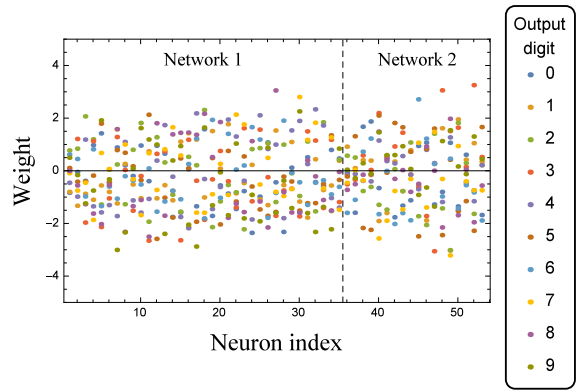}
\caption{Weights of the $[784,48,35,10]+[784,24,18,10]$ PNN when optimized at
epoch $89$ of Fig.7. The $x$-coordinate denotes the index of the neurons in
the last hidden layer, where neurons $1\symbol{126}35$ ($36\symbol{126}53$)
belong to Network 1 (Network 2), i.e., the $[784,48,35,10]$ ($[784,24,18,10]$%
) sub-network. The $y$-coordinate denotes the weights of the outputs of
these neurons when calculating the weighted inputs to the output neurons.
Different colors denote different output neurons that lead to different
output digits. For example, the orange dot at coordinates $(52,3.25)$\ (the
highest point on the top-right of the figure) means that for the output
neuron corresponding to the output digit $3$, the output of the $52$nd
neuron in the last hidden layer has the weight $3.25$.}
\label{fig:epsart}
\end{figure}


This experiment reveals that in a PNN, the contribution of an inferior
sub-network could be comparable with that of the other sub-network. This is
yet another property that differs from natural sense organs. For example, the
two eyes of many people (me included) have visual acuity differences. Most
of the time I find myself mainly relying on the better eye only, while the
worse one contributes little for my vision. But the current experiment shows
that this is not always the case for artificial neural networks.

\bigskip

\textit{(3) Disagreed results.}


\begin{table*}[htbp]
\caption{The numbers of different types of correct results of each PNN. The
total number of the evaluation data is $10000$. The activation function of
each neuron is chosen as the sigmoid function. Field name explanation: \\
Training method A: The two sub-networks in a PNN are trained separately for $%
60$ epochs, then connected in parallel and trained together for another $40$
epochs. \\
Training method B: The two sub-networks are connected from the very start
and trained for $100$ epochs. \\
Total: The total number of results that the PNN classifies correctly. \\
Type I: The number of results when both sub-networks make the right
classification. \\
Type II: The number of results when only Network 1 is wrong. \\
Type III: The number of results when only Network 2 is wrong. \\
Type IV: The number of results when both sub-networks are wrong but the
whole PNN is right.}
\label{tab:1}       
\begin{tabular}{c|c|c|c|c|c|c}

\hline

\multirow{2}[3]{*}{PNN} & \multirow{2}[3]{*}{Training method} & \multicolumn{5}{c}{Number of results} \\
\cline{3-7}          &       & Total & Type I & Type II & Type III & Type IV \\

\hline

$[784,48,35,10]+[784,50,10]$ & A     & 9805  & 7044  & 59    & 2610  & 92 \\
$[784,48,35,10]+[784,50,10]$ & B     & 9777  & 118   & 8055  & 864   & 740 \\
$[784,48,35,10]+[784,48,35,10]$ & A     & 9817  & 9244  & 340   & 211   & 22 \\
$[784,48,35,10]+[784,48,35,10]$ & B     & 9816  & 3138  & 2341  & 3812  & 525 \\
$[784,48,35,10]+[784,24,18,10]$ & B     & 9802  & 3222  & 1141  & 4262  & 1177 \\
$[784,45,79,10]+[784,50,10]$ & B     & 9776  & 0     & 8369  & 963   & 444 \\

\hline

\end{tabular}
\end{table*}


The result in (2) shows that a sub-network could still play an important
role in the whole PNN even if its own classification accuracy $\alpha
_{i}^{\prime }$ is low. Then what is this sub-network busy working on? To
answer this question, recall that at the end of (1.1), we said that in a
PNN, the sub-networks are turned to capture other features of the input. Now
we elaborate this point with more data analysis.

For each handwritten digit image among the MNIST data set being input to the
PNN as an evaluation data, let $y$ denote the actual digit that it stands
for, $r_{1}$, $r_{2}$\ and $r_{p}$\ denote the classification results of
Network 1, Network 2 and the whole PNN, respectively, where $r_{1}$ and $%
r_{2}$\ are calculated from Eq. (\ref{NNp}). (We also tried calculating
using Eq. (\ref{NN}). The qualitative results below remain the same despite
that the exact values will vary.) All the results that the PNN classifies
correctly can be categorized as four types:

Type I: $r_{p}=y$\ and $r_{1}=y$\ and $r_{2}=y$.

Type II: $r_{p}=y$\ and $r_{1}\neq y$\ and $r_{2}=y$.

Type III: $r_{p}=y$\ and $r_{1}=y$\ and $r_{2}\neq y$.

Type IV: $r_{p}=y$\ and $r_{1}\neq y$\ and $r_{2}\neq y$.

The first five rows (without counting the header) of Table I list the
number of each type of the results of the PNNs studied in (1) and (2), whose
biases and weights take the values when $\alpha _{para}$ is optimized. For
example, the first row is corresponding to the $[784,48,35,10]+[784,50,10]$
PNN when its $\alpha _{para}$ is optimized at epoch $86$ in experiment (1.1).

What really catches our eyes is that all PNNs have a non-trivial amount of
type IV results, i.e., the results $r_{1}$, $r_{2}$ of both sub-networks
disagree with the actual digit $y$, still the whole PNN gives the correct
result $r_{p}$. This result once again reveals that the sub-networks
collaborate in a different way from our natural sense organs. Try close your
right eye and read the following text with your left eye. Now open your
right eye and close your left eye and read it again. Surely your both eyes
recognize the same text correctly every time. While we still do not know
exactly what happens in our brain during this process, it is unlikely
that our left eye sees X, right eye thinks it as Y, while our brain comes up
with Z and it is correct! But the existence of type IV results reveals that
artificial neural networks can indeed work this way.

Let us look at the type IV results even closer. Table II shows a portion of
the $92$ type IV results mentioned in the first row of Table I (the full list
is also available, see \textit{Data availability} section). We can see that $r_{p}$\
is not a single-valued function of $r_{1}$ and $r_{2}$. For example, when $%
r_{1}=2$ and $r_{2}=5$, $r_{p}$\ can either be $1$, $7$ or $8$, which cannot be deduced from $r_{1}$ and $r_{2}$ uniquely. Thus, when
serving as a part of the PNN for the classification task, it is clear that
the actual task of each sub-network is not to provide its own classification
result $r_{i}$ alone. Instead, it is to provide other information of the
handwritten digit image via the distribution of its weighted input to all
output neurons. Then the whole PNN gathers all these different informations
from every sub-network and draws its final conclusion.


\begin{table}[b]
  \centering
  \caption{Some of the type IV results of the $[784,48,35,10]+[784,50,10]$ PNN
optimized in experiment (1.1). $r_{1}$, $r_{2}$\ and $r_{p}$\ denote the
classification results of Network 1, Network 2 and the whole PNN,
respectively.}
\label{tab:2}       
    \begin{tabular}{l|r|r|r|r|r|r|r|r|r|r|r|r|r|r|r|r|r|r|r|r|r|r|r|r}
    \hline
    $r_{1}$    & 0     & 0     & 2     & 2     & 2     & 2     & 2     & 2     & 2     & 2     & 2     & 2     & 2     & 2     & 3     & 3     & 3     & 3     & 3     & 3     & 3     & 3     & 3     & 3 \\
    \hline
    $r_{2}$    & 5     & 5     & 0     & 0     & 2     & 2     & 3     & 5     & 5     & 5     & 5     & 5     & 5     & 5     & 0     & 0     & 2     & 2     & 5     & 5     & 5     & 5     & 5     & 5 \\
    \hline
    $r_{p}$    & 4     & 8     & 4     & 7     & 4     & 4     & 4     & 1     & 7     & 7     & 8     & 8     & 8     & 8     & 8     & 8     & 8     & 8     & 8     & 8     & 8     & 8     & 8     & 9 \\
    \hline
    \end{tabular}%
  \label{tab:addlabel}%
\end{table}%


This is similar to the networks with pooling \cite{ml100}, where certain parts of the network take charge of capturing certain features of the input. But as we mentioned, in these networks each neuron in the pooling layer connects with a portion of the neurons in the previous layer only. Now our result shows that the division of labor also takes place even when every sub-network connects with all input neurons.

We would like to mention another interesting observation that we found when
studying the $[784,45,79,10]+[784,50,10]$ PNN.
We optimized this PNN with the training method B and obtained $\max (\alpha
_{para})=97.76\%$ at epoch $91$. The last row of Table I shows the numbers
of different types of correct results at this stage. Surprisingly, the
number of type I results turns out to be $0$. That is, the two sub-networks
never reach the correct classification result simultaneously, while the
whole PNN still comes up with a very high classification accuracy. This
seems to be another divergence from the property of natural sense organs. We
repeated the training of this PNN three times and the absence of type I
results occurs twice (see \textit{Data availability} section for full data),
therefore it does not seem to be a rare case.

\textit{(4) Using other activation functions.}

The above experiments are all performed with the sigmoid function serving as
the activation function. Now we try these PNNs by replacing it with the
rectified linear unit (ReLU) function%
\begin{equation}
f(z)=\max (0,z)
\end{equation}%
or the Tanh function%
\begin{equation}
f(z)=\frac{e^{z}-e^{-z}}{e^{z}+e^{-z}}.
\end{equation}

Like experiment (1.4), with each form of the activation functions, we
trained the $[784,48,35,10]+[784,48,35,10]$ PNN using training method B for $%
9$ trials. The number of type IV results ($n_{IV}$) and $\max
(\alpha _{para})$\ of each trial are listed in Table III.


\begin{table}[b]
  \centering
  \caption{Comparison between three $[784,48,35,10]+[784,48,35,10]$ PNNs, which
use the sigmoid, ReLU, or Tanh function as the activation function,
respectively. Each PNNs are trained for $9$ trials using training method B. $%
n_{IV}$ is the number of type IV results, and $\alpha $ is
the classification accuracy $\max (\alpha _{para})$ of the PNN when it is optimized.}
\label{tab:3}       
    \begin{tabular}{c|c|c|c|c|c|c}
    \hline
    \multirow{2}[3]{*}{Trial} & \multicolumn{2}{c|}{sigmoid} & \multicolumn{2}{c|}{ReLU} & \multicolumn{2}{c}{Tanh} \\
\cline{2-7}          & $n_{IV}$   & $\alpha $(\%) & $n_{IV}$   & $\alpha $(\%) & $n_{IV}$   & $\alpha $(\%) \\
    \hline
    1     & 525   & 98.16 & 338   & 97.84  & 1056  & 97.72  \\
    2     & 684   & 98.04 & 1028  & 97.90  & 504   & 97.65  \\
    3     & 384   & 98.21 & 112   & 97.80  & 695   & 97.68  \\
    4     & 2086  & 98.16 & 950   & 97.91  & 629   & 97.75  \\
    5     & 256   & 98.08 & 224   & 97.84  & 504   & 97.90  \\
    6     & 284   & 98.13 & 393   & 97.65  & 364   & 97.86  \\
    7     & 234   & 98.18 & 638   & 97.81  & 1911  & 97.58  \\
    8     & 1056  & 98.19 & 175   & 97.83  & 515   & 97.91  \\
    9     & 286   & 98.18 & 259   & 97.92  & 1392  & 97.72  \\
    \hline
    Average & 644   & 98.15 & 457   & 97.83 & 841   & 97.75 \\
    \hline
    \end{tabular}%
  \label{tab:addlabel}%
\end{table}%


From the average of $n_{IV}$, we can see that although the PNN with ReLU
function still shows divergence from the connection of natural organs, it is
the closest among the three, since it has the lowest average $\bar{n}%
_{IV}=457$. The PNN with sigmoid function ($\bar{n}_{IV}=644$) could have
been close, if not for the crazy high value $n_{IV}=2086$ in Trial 4 (the
average of the other 8 trials will drop down to $464$). With the Tanh
function, the result $\bar{n}_{IV}=841$ is undoubtedly the highest. Even if
we discard the highest value $n_{IV}=1911$ in Trial 7, the average of the
rest trials is still $707$.

\section*{Discussion}

In summary, when two sub-networks connect and form an PNN, our first three
experiments show that there are three properties which distinguish artificial
neural networks from the parallel connection of natural biological sense
organs. Namely, experiment (1) shows that when the classification accuracy $%
\alpha _{para}$\ of the PNN is optimized, the classification accuracies $%
\alpha _{i}$\ and $\alpha _{i}^{\prime }$\ of each sub-network will be lower
than the maximal values that could be obtained when being trained and
optimized alone. Even if the sub-networks are trained and optimized
separately beforehand, $\alpha _{i}$\ and $\alpha _{i}^{\prime }$\ will drop
once they are connected. But for natural organs such as two eyes, connecting to
the same brain should not make the eye sight of each of them deteriorated.
Training two eyes separately does not seem to be a common practice either.
Also, experiment (2) shows that the contribution of an inferior sub-network
to the whole PNN is on par with that of a superior sub-network. This is in
contrast to two real eyes with visual acuity differences, where the better
eye generally does most of the job. Moreover, experiment (3) shows that the
PNN can output the correct result when both sub-networks give incorrect and
disagreed results, i.e., the type IV results, which is also impossible for
natural organs like we said in the introduction.

As an amusing thought, assume that someday we could build a robot with these
artificial neural networks serving as sense organs. If these artificial
organs are connected to the cyberbrain ever since the robot is
\textquotedblleft born\textquotedblright , then weird things happen as they
\textquotedblleft grow\textquotedblright . Like our experiments (1.2), (1.4)
and (3), as they are trained together, they will eventually turn into fuzzy
eyes, mishearing ears and malfunctioning nose, still the cyberbrain becomes
more and more clever! While this may somehow like a wise old man, it does not appear to be the meaning of \textquotedblleft bionics\textquotedblright that people agree with. On the
other hand, experiments (1.3) suggested that the final result will look
closer to natural organs if the artificial ones are trained and optimized
separately before putting them together. Therefore, in the future if we want a robot that could act like human, then all parts have to be trained separately before being assembled. We could not expect them to be put together at
\textquotedblleft birth\textquotedblright\ and then grow and learn as an entirety like a human baby.

Back to the topic, these three experiments lead us to the idea that the
three divergences from natural sense organs, especially the number (or ratio)
of type IV results, can be used as a criterion for measuring the bionic
level of an artificial neural network. The more serious divergences we
observe, the less likely it is for such a network to mimic the actual
internal mechanism of real living creatures.

Experiment (4) further proves that this criterion can indeed manifest the
differences among neural networks when other parameters fail. As shown in
Table III, when the activation function of an artificial neural network is
chosen as the sigmoid, ReLU, or Tanh functions, respectively, there is no
significant differences on their classification accuracies. But when we look
at their numbers of type IV results, the ReLU function shows the potential to
make the neural network more bionic according to our criterion.

This may somewhat deepen our understanding on the ReLU (rectified linear
unit) function. As pointed out in chapter 3 of Ref. \cite{ml98},
\textquotedblleft some recent work on image recognition has found
considerable benefit in using rectified linear units through much of the
network. However, ..., we do not yet have a really deep understanding of
when, exactly, rectified linear units are preferable, nor
why.\textquotedblright\ Also in chapter 6, \textquotedblleft a few people
tried rectified linear units, often on the basis of hunches or heuristic
arguments. ... In an ideal world we'd have a theory telling us which
activation function to pick for which application. But at present we're a
long way from such a world. ... Today, we still have to rely on poorly
understood rules of thumb and experience.\textquotedblright\ But now we find
that even if there is no obvious difference between ReLU and other functions
on the classification accuracy or other properties, still we can test
their bionic level using our criterion. And the observation that ReLU has
benefit in tasks like image recognition may indicate that it is a better
approximation of the input-output dependence of a real living neuron than other functions do.

If this is indeed the case, then it can further lead to a method for learning the input-output dependence of natural neurons, which is difficult to measure directly with biological experiments. That is, we can try proposing another form of the activation function, then build artificial networks upon it and test their bionic level with our criterion. If they appear to be more bionic than the networks using ReLU function, then we know that the curve of this activation function describes the input-output dependence of natural neurons even better.

Note that our criterion does not indicate that one neural network is superior or
inferior to another. It merely tells which one is more bionic.

Finally, the activation function is only one of the characters that can
affect the bionic level of an artificial neural network. Other characters
(e.g., the number of layers, the usage of pooling, etc.) may also show their
impact if we study the number of type IV results. These works could be
considered for future researches.

\bigskip

\section*{Methods}

Our PNNs were trained using the Python programs accompany with this paper
(see Code availability section). The programs are based on the program
network2.py supplied with Ref. \cite{ml98}, where the training methods
include L2 regularization, the usage of the cross-entropy cost function, randomly
shuffling the training data, and the weights input to a neuron are
initialized as Gaussian random variables with mean $0$ and standard
deviation $1$ divided by the square root of the number of connections input
to the neuron, as elaborated in chapter 3 of Ref. \cite{ml98}. Note that our
purpose is to study the behaviors of parallel-connected networks, instead of
trying to find the highest possible classification accuracy. Thus, to keep
things simple, we did not use other advanced training techniques, e.g.,
dropout, artificial expansion of the training date, softmax, local receptive
fields and pooling, etc.

Due to the randomness in the initialization of the biases and weights and
the shuffling of the training data, the exact values of the classification
accuracies in different runs of the programs will vary. To avoid our results
being too extreme, the training of each PNN studied in the \textit{%
Experiments} section was repeated 3 times (9 times for experiments (1.4) and
(4)). Then we picked the one whose $\max (\alpha _{para})$ takes the middle
value of all runs, and labelled it as Trial 1. The other runs were labelled as
Trials 2\symbol{126}3 (2\symbol{126}9). We used the data in
Trial 1 to pot the figures in the \textit{Experiments} section. The data in
Trials 2\symbol{126}3 (2\symbol{126}9) are not shown. But we also provide
them with the paper for reference (see \textit{Data availability} section).

\bigskip

\textbf{Data availability}

The raw experimental data of this study are publicly available at
https://github.com/gphehub/pnn2204.

\textbf{Code availability}

The software codes for generating the experimental data are publicly
available at https://github.com/gphehub/pnn2204.

\bigskip

\textbf{Acknowledgements}

The work was supported in part by Guangdong Basic and Applied Basic Research
Foundation under Grant No. 2019A1515011048.

\textbf{Author contributions}

G.P.H. researched and wrote this paper solely.

\textbf{Competing interests}

The author declares no competing interests.

\end{document}